\documentclass[runningheads,a4paper]{llncs}

\usepackage[english]{babel}
\usepackage{graphicx}
\usepackage[colorinlistoftodos]{todonotes}

\title{Multi-Target Regression via \\ Random Linear Target Combinations}
\author{Grigorios Tsoumakas \and Eleftherios Spyromitros-Xioufis \and \\ Aikaterini Vrekou \and Ioannis Vlahavas}

\authorrunning{G. Tsoumakas \and E. Spyromitros-Xioufis \and A. Vrekou  \and I. Vlahavas}

\institute{Department of Informatics, Aristotle University of Thessaloniki,\\ 54124 Thessaloniki, Greece \\ \email{greg,espyromi,agvrekou,vlahavas@csd.auth.gr}}

\begin{document}
\mainmatter

\maketitle

\begin{abstract}
Multi-target regression is concerned with the simultaneous prediction of multiple continuous target variables based on the same set of input variables. It arises in several interesting industrial and environmental application domains, such as ecological modelling and energy forecasting. This paper presents an ensemble method for multi-target regression that constructs new target variables via random linear combinations of existing targets. We discuss the connection of our approach with multi-label classification algorithms, in particular RA$k$EL, which originally inspired this work, and a family of recent multi-label classification algorithms that involve output coding. Experimental results on 12 multi-target datasets show that it performs significantly better than a strong baseline that learns a single model for each target using gradient boosting and compares favourably to multi-objective random forest approach, which is a state-of-the-art approach. The experiments further show that our approach improves more when stronger unconditional dependencies exist among the targets.
\keywords{multi-target regression, multi-output regression, multivariate regression, multi-label classification, output coding, random linear combinations}

\end{abstract}

\section{Introduction}

Multi-target regression, also known as multivariate or multi-output regression, aims at simultaneously predicting multiple continuous target variables based on the same set of input variables. Such a learning task arises in several interesting application domains, such as predicting the wind noise of vehicle components \cite{kuznar:2009}, ecological modelling \cite{kocev:2009}, water quality monitoring \cite{dzeroski:2000}, forest monitoring \cite{dzeroski+etal:2006} and more recently energy-related forecasting\footnote{http://www.gefcom.org}, such as wind and solar energy production forecasting and load/price forecasting.

Multi-target regression can be considered as a sibling of multi-label classification \cite{tsoumakas+etal:2010b,zhang+zhou:2013}, the latter dealing with multiple {\em binary} target variables, instead of {\em continuous} ones. Recent work \cite{spyromitros:2014:arxiv} stressed the close connection among these two tasks and argued that ideas from the more popular and developed area of multi-label learning could potentially be transferred to multi-target regression. Following up this argument, we present here a multi-target regression algorithm that was conceived as analogous to the RA$k$EL \cite{tsoumakas+etal:2011} multi-label classication algorithm. In particular, the proposed method creates new target variables by considering random linear combinations of $k$ original target variables. Experiments on 12 multi-target datasets show that our approach is significantly better than a strong baseline that learns a single model for each target using gradient boosting \cite{friedman:2001} and compares favourably to the state-of-the-art multi-objective random forest approach\cite{kocev:2007}. The experiments further show that our approach improves more when stronger unconditional dependencies exist among the targets.

The rest of this paper is organized as follows. Section \ref{sec:rw} discusses related work on multi-target regression, as well as on output coding, a family of multi-label learning algorithm of similar nature to our approach, which is presented in Section \ref{sec:rltc}. Section \ref{sec:setup} presents the setup of our empirical study (methods and their parameters, implementation details, evaluation process, datasets) and Section \ref{sec:results} discusses our experimental results. Finally, section \ref{sec:conclusions} summarizes the conclusions of this work and points to future work directions.

\section{Related Work}
\label{sec:rw}

\subsection{Multi-Target Regression}

Multivariate regression was studied many years ago by statisticians and two of the earliest methods were reduced-rank regression \cite{izenman:1975} and C\&W \cite{breiman:1997}. A large number of methods for multi-target regression are derived from the predictive clustering tree (PCT) framework \cite{blockeel:1998}. These are presented in more detail in subsequent paragraphs. An approach for learning multi-target model trees was proposed in \cite{appice:2007}. One can also find methods that deal with multi-target regression problems in the literature of the related topics of transfer learning \cite{piccart+etal:2008} and multi-task learning \cite{jalali+etal:2010}. Undoubtedly, the simplest approach to multi-target regression is to independently construct one regression model for each  of the target variables.

The main difference between the PCT algorithm and a standard decision tree is that the variance and the prototype functions are treated as parameters that can be instantiated to fit the given learning task. Such an instantiation for multi-target prediction tasks are the multi-objective decision trees (MODTs), where the variance function is computed as the sum of the variances of the targets, and the prototype function is the vector mean of the target vectors of the training examples falling in each leaf \cite{blockeel:1998,blockeel:1999}. Bagging and random forest ensembles of MODTs were developed in \cite{kocev:2007} and found significantly more accurate than MODTs and equally good or better than ensembles of single-objective decision trees for both regression and classification tasks. In particular, multi-objective random forest (MORF) yielded better performance than multi-objective bagging.

Motivated by the interpretability of rule learning algorithms, other researchers developed multi-target rule learning algorithms that again fall in the PCT framework. Focusing on multi-label classification problems, \cite{zenko:2008} proposed the predictive clustering rules (PCR) method that extends the PCT framework by combining a rule learning algorithm with a search heuristic that derives from clustering. PCR yielded comparable accuracy to using multiple single-target rule learners using a much smaller and interpretable collection of rules. Later, the FIRE rule ensemble algorithm \cite{aho+etal:2009} was proposed, specifically designed for multi-target regression. FIRE  works by first transforming an ensemble of decision trees into a collection of rules and then using an optimization procedure that assigns proper weights to individual rules in order to prune the initial rule set without compromising its accuracy. The connection of this method to the PCT framework lies in the fact that the ensemble of trees comes from the MORF method of \cite{kocev:2007}. Recently, \cite{aho+etal:2012} presented FIRE++, an improved version of FIRE, which among other optimizations, offers the ability to combine rules with simple linear functions. FIRE++ was found better than FIRE, but slightly worse than the less interpretable MORF.

\subsection{Output Coding}
\label{sec:output-coding}

Linear combinations of targets have been recently used by a number of output coding approaches \cite{hsu+etal:2009,zhang+schneider:2011,zhang+schneider:2012,tai+lin:2012} for the related task of multi-label classification \cite{tsoumakas+etal:2010b,zhang+zhou:2013}. The motivation of the methods in \cite{hsu+etal:2009} and \cite{tai+lin:2012} was the reduction of large output spaces for improving computational complexity, which goes towards the opposite direction of our approach. The methods in \cite{zhang+schneider:2011} and \cite{zhang+schneider:2012} on the other hand, aimed at improving the prediction accuracy similarly to our approach.

The approach most similar to ours is the chronologically first one \cite{hsu+etal:2009}, which is based on the technique of compressed sensing and consideres random linear combinations of the labels. This is also the only output coding method from the ones mentioned here, where the dimensionality of the new output space is allowed to be larger than the original output space, as in our case. Besides the opposite motivation (compression of output space) compared to our approach, \cite{hsu+etal:2009} starts from the concept of output sparsity (sparsity of the output conditioned on the input), while in multi-target data, the output space is generally non-sparse. The encoding step of \cite{hsu+etal:2009} is therefore based on compression matrices that satisfy a restricted isometry property, based on a sparsity level defined by the user and the decoding step is based on sparse approximation algorithms. In contrast, our approach uses uniform non-zero random weights for a user-defined number of targets in the encoding step, and standard unregularized least squares in the decoding step.

\section{Random Linear Target Combinations}
\label{sec:rltc}

Consider a set of $p$ input variables $\mathbf{x} \in R^p$ and a set of $q$ target variables $\mathbf{y} \in \mathcal{R}^q$. We have a set of $m$ training examples: $\mathbf{D} = (\mathbf{X}, \mathbf{Y}) = \{(\mathbf{x^{(i)}}, \mathbf{y^{(i)}})\}_{i=1}^{m}$, where $\mathbf{X}$ and $\mathbf{Y}$ are matrices of size $m \times p$ and $m \times q$, respectively.

Our approach constructs $r >> q$ new target variables via corresponding random linear combinations of $\mathbf{y}$. To achieve this, we define a coefficient matrix $\mathbf{C}$ of size $q \times r$ filled with random values uniformly chosen from $[0..1]$. Each column of this matrix contains the coefficients of a linear combination of the target variables. Multiplying $\mathbf{Y}$ with $\mathbf{C}$ leads to a transformed multi-target training set $\mathbf{D'} = (\mathbf{X}, \mathbf{Z})$, where $\mathbf{Z}=\mathbf{Y}\mathbf{C}$ is a matrix of size $m \times r$ with the values of the new target variables. A user-specified multi-target regression learning algorithm is then applied to $\mathbf{D'}$ in order to build a corresponding model.

Note that our approach expects that the original target variables take values from the same domain, as otherwise their linear combinations could be dominated by the values of targets with a much wider domain than the others. To ensure this, it applies 0-1 normalization in order to bring the values of all targets into the range [0..1].

We consider an additional parameter $k \in \{2,\ldots,q\}$ for specifying the number of original target variables involved in each random linear combination, by setting the coefficients for the rest of the target variables to zero. Higher $k$ means that potential correlations among more targets are being considered. However, at the same time, it means that the new targets are more difficult to predict, especially in the absence of actual correlations among the targets. We therefore hypothesize that low $k$ values will lead to the best results. In practice, when $k < q$, for each linear combination our approach selects $k$ targets at random, but with priority to targets with the lowest frequency of participation to previously considered linear combinations. This ensures that all targets will participate in $\mathbf{C}$ as equivallently (i.e. with similar frequency) as possible.

Given a new test instance, $\mathbf{x'}$, the multi-target regression model is first invoked to obtain a vector $\mathbf{z'}$ with $r$ predictions. The estimates $\hat{\mathbf{y}}'$ for the original target variables are then obtained by solving for $\hat{\mathbf{y}}'$ the following overdetermined (as $r >> q$) system of linear equations: $\mathbf{C}^{\top}\hat{\mathbf{y}}'=\mathbf{z'}$.

\begin{figure}
\centering
\includegraphics[width=\textwidth]{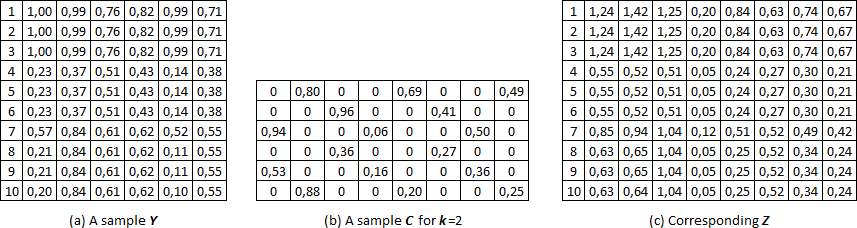}
\caption{\label{fig:example}An example of our approach. The $q=6$ targets of a multi-target regression dataset with $m=10$ examples is shown in (a). A coefficient matrix for $k=2$ and $r=8$ is shown in (b). The values of the new targets is shown in (c).}
\end{figure}

As an example of our approach, consider a multi-target training set with $q=6$ targets and $m=10$ training examples. Figure \ref{fig:example}(a) shows the normalized targets, $\mathbf{Y}$ of such a dataset, based on the first 10 training examples of the atp1d dataset (see Section \ref{sec:datasets} for a description of this dataset). Figure \ref{fig:example}(b) shows a potential coefficient matrix $\mathbf{C}$ for $r=8$ and $k=2$. Finally, Figure \ref{fig:example}(c) shows the values of the new targets $\mathbf{Z}$.

Our approach was inspired from recent work on drawing parallels between multi-label classification and multi-target regression \cite{spyromitros:2014:arxiv} and conceived as the twin of the multi-label classification algorithm RA$k$EL \cite{tsoumakas+etal:2011} for multi-target regression tasks. Similarly to RA$k$EL, our approach aims to exploit correlations among target variables on one hand and to achieve the error-correction effect of ensemble methods on the other hand, as it implicitly pools multiple estimates for each original target variable (one for each linear combination that it participates in). We therefore expect that the larger $r$ is, the better the estimate of the original target variables. Our approach follows the {\em randomness injection} paradigm of ensemble construction \cite{dietterich:2000} at a larger degree than RA$k$EL, as it may combine the same target variables twice, but with different random coefficients. Randomness is a key component for improving supervised learning methods \cite{breiman:2001,hinton+etal:2012}.

After inventing our approach, we realized that linear target combination approaches have been used for multi-label data in the past. From this viewpoint, our approach could also be considered as a sibling of multi-label compressed sensing \cite{hsu+etal:2009}, if we set aside the different goal and the technical differences among the two approaches discussed in Section~\ref{sec:output-coding}.

\section{Experimental Setup}
\label{sec:setup}

This section offers details on the setup of the experiments that we conducted. We first present the participating methods and their parameters, then provide implementation details, followed by a description of the evaluation measure and process that was followed. We conclude this section by presenting the datasets that were used, their main statistics, as well as statistics of the pairwise correlations among their target variables.

\subsection{Methods and Parameters}

Our approach (dubbed RLC) is parameterized by the number of new target variables, $r$, the number of original target variables to combine, $k$, the multi-target regression algorithm that is used to learn from the transformed multi-target training set $\mathbf{D'}$ and the approach used to solve the overdetermined system of linear equations during prediction. The first two we discuss together with the results in Section \ref{sec:results}. The multi-target regression algorithm we employ is to learn a single independent regression model for each target (dubbed ST). Each regression model is built using gradient boosting \cite{friedman:2001} with a 4-terminal node regression tree as the base learner, a learning rate of 0.1 and 100 boosting iterations. The system of linear equations is solved by the unregularized least squares approach.

The multi-target regression algorithm employed by our approach, ST with gradient boosting, is also directly used on the original target variables as a baseline. We further compare our approach against the state-of-the-art multi-objective random forest algorithm \cite{kocev:2007} (dubbed MORF). We used an ensemble size of 100 trees and the values suggested in \cite{kocev:2007} for the rest of the parameters.

\subsection{Implementation}

The proposed method was implemented within the open-source multi-label learning Java library Mulan\footnote{\url{http://mulan.sourceforge.net}}
\cite{tsoumakas+etal:2011b}, which has been recently expanded to handle multi-target prediction tasks and includes an implementation of ST too, as well as a wrapper of the CLUS software\footnote{http://dtai.cs.kuleuven.be/clus/}, including support for MORF. Mulan is built on top of Weka\footnote{\url{http://www.cs.waikato.ac.nz/ml/weka}}
\cite{hall+etal:2009}, which includes an implementation of gradient boosting. Therefore, the comparative evaluation of all methods was achieved using a single Java-based software framework.

In support of open science, Mulan includes a package called {\em experiments}, which contains experimental setups of various algorithms based on the corresponding papers. To ease replication of the experimental results of this paper, we have included a class called {\em ExperimentRLC} in that package.

\subsection{Evaluation}

We use the average Relative Root Mean Squared Error (aRRMSE) as evaluation measure. The RRMSE for a target is equal to the  Root Mean Squared Error (RMSE) for that target divided by the RMSE of predicting the average value of that target in the training set. This standardization facilitates performance averaging across non-homogeneous targets.

The aRRMSE of a multi-target model $h$ that has been induced from a train set $\mathbf{D_{train}}$ is estimated based on a test set $\mathbf{D_{test}}$ according to the following equation:
\[
aRRMSE(h,\mathbf{D_{test}}) = \frac{1}{q} \sum_{j=1}^{q}RRMSE
= \frac{1}{q} \sum_{j=1}^{q}\sqrt{
    \frac
        {
            \sum_{(\mathbf{x},\mathbf{y}) \in \mathbf{D_{test}}} (h(\mathbf{x})_{j}-y_{j})^{2}
        }
        {
           \sum_{(\mathbf{x},\mathbf{y}) \in \mathbf{D_{test}}} (\bar{y}_{j}-y_{j})^{2}
        }
  }
\]
where $\bar{y}_{j}$ is the mean value of target variable $y_{j}$ within  $\mathbf{D_{train}}$ and $h(\mathbf{x})_{j}$ is the output of $h$ for target variable $y_{j}$.

The aRRMSE measure is estimated using the hold-out approach for large datasets, while 10-fold cross-validation is employed for small datasets.

\subsection{Datasets}
\label{sec:datasets}

Our experiments are based on 12 datasets\footnote{\url{http://users.auth.gr/espyromi/datasets.html}}. Table~\ref{tab:statistics} reports the name (1st column), abbreviation (2nd column) and source (3rd column) of these datasets, the number of instances of the train and test sets or the total number of instances if cross-validation was used (4th column), the number, $p$, of input variables (5th column) and the number, $q$, of output variables (6th column).

\begin{table}
\caption{Name, abbreviation, source, number of train and test examples or total number of examples in the case of cross-validation, number of input variables and number of output variables per dataset used in our empirical study.}
\centering
\begin{tabular}{cccccc}
\hline
Name & Abbreviation & Source & Examples & $p$ & $q$ \\
\hline
Airline Ticket Price 1 & atp1d & \cite{spyromitros:2014:arxiv} & 337 & 411 & 6 \\
Airline Ticket Price 2 & atp7d & \cite{spyromitros:2014:arxiv} & 296 & 411 & 6 \\
Electrical Discharge Machining & edm & \cite{karalic+bratko:1997} & 154 & 16 & 2\\
Occupational Employment Survey 1 & oes1997 & \cite{spyromitros:2014:arxiv} & 334 & 263 & 16 \\
Occupational Employment Survey 2 & oes2010 & \cite{spyromitros:2014:arxiv} & 403 & 298 & 16 \\
River Flow 1 & rf1 & \cite{spyromitros:2014:arxiv} & 4165/5065 & 64 & 8 \\
River Flow 2 & rf2 & \cite{spyromitros:2014:arxiv} & 4165/5065 & 576 & 8 \\
Solar Flare 1 & sf1969 & \cite{asuncion+newman:2007} & 323 & 26 & 3 \\
Solar Flare 2 & sf1978 & \cite{asuncion+newman:2007} & 1066 & 27 & 3 \\
Supply Chain Management 1 & scm1d & \cite{spyromitros:2014:arxiv} & 8145/1658 & 280 & 16 \\
Supply Chain Management 2 & scm20d & \cite{spyromitros:2014:arxiv} & 7463/1503 & 61 & 16 \\
Water Quality & wq & \cite{dzeroski:2000} & 1060 & 16 & 14 \\
\hline
\end{tabular}
\label{tab:statistics}
\end{table}

\begin{figure}
\centering

\begin{minipage}{0.49\textwidth}
  \centering
  \includegraphics[width=\linewidth]{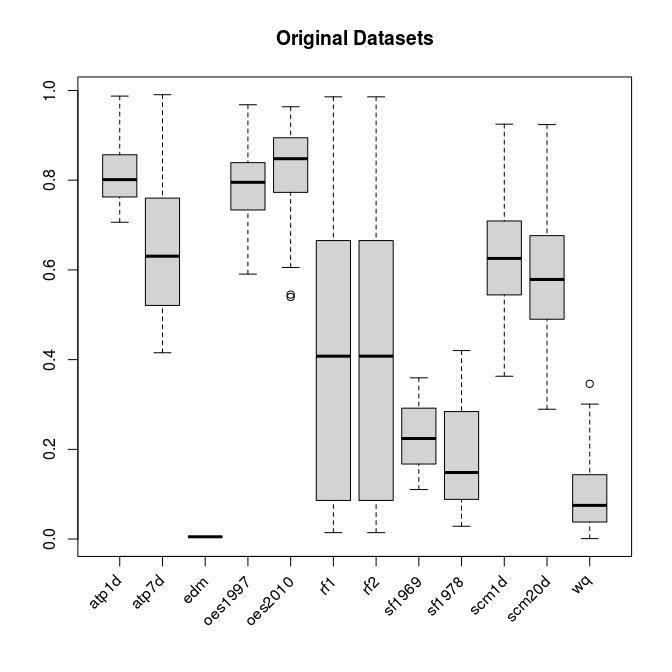}
  \caption{\label{fig:boxplots_or}Box-plots summarizing the distribution of all pairwise target correlations for all datasets.}
\end{minipage}
\hspace{15 pt}
\begin{minipage}{0.45\textwidth}
  \centering
  \includegraphics[width=\linewidth]{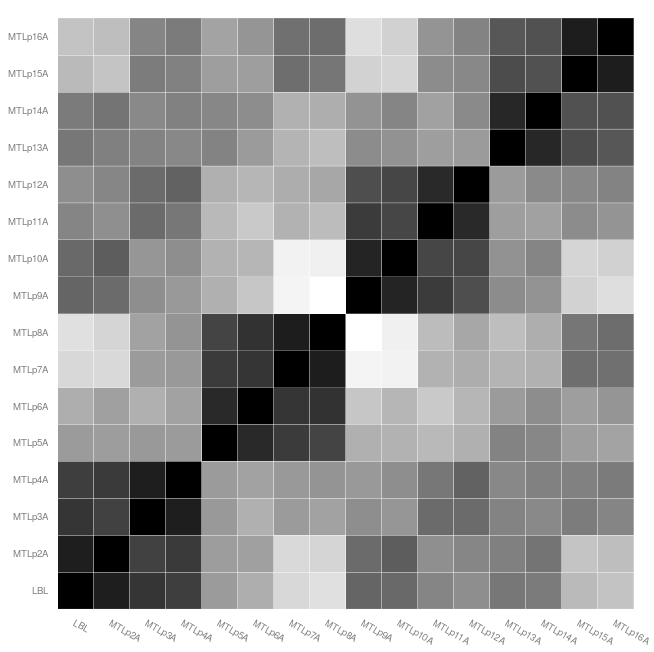}
  \caption{\label{fig:scm20d_heatmap}Heat-map of the pairwise target correlations for the {\em scm20d} dataset.}
\end{minipage}
\end{figure}

One of the motivations of our approach is the exploitation of potential dependencies among the targets. We hypothesize that our approach will do better in datasets where target dependencies exist. To facilitate the discussion of results in this context, Figure~\ref{fig:boxplots_or} shows box-plots summarizing the distribution of the correlations among all pairs of targets for all datasets, while Figure~\ref{fig:scm20d_heatmap} shows a heat-map of the pairwise target correlations for a sample dataset with a relatively large number of targets ({\em scm20d}). The rest of this section provides a short description for each of the datasets.

\subsubsection{Airline Ticket Price}

The {\em airline ticket price} dataset \cite{spyromitros:2014:arxiv} was constructed for the prediction of airline ticket prices for a specific departure date. There are two versions of this datasets. The target attributes are the next day price (atp1d) or the minimum price within the next 7 days (atp7d) for 6 characteristics: any airline with any number of stops, any airline non-stop only, Delta Airlines, Continental Airlines, Airtran Airlines and United Airlines. The input attributes are the number of days between the observation and departure date, 7 binary attributes that refer to the day-of-the-week of the observation date and the complete enumeration of: 1) the minimum price, mean price and number of quotes from, 2) all airlines and from each airline quoting more than 50\% of the observation days, 3) for non-stop, one-stop and two-stop flights, 4) for the current day, previous day and two days before. There are 411 input attributes in total.

\subsubsection{Electrical Discharge Machining}

The {\em electrical discharge machining} dataset \cite{karalic+bratko:1997} represents a two-target regression
problem. The task is to shorten the machining time by reproducing the behavior of a human operator which controls the values of two variables. Each of the target variables takes 3 distinct numeric values (1,0,1) and there are 16 continuous input variables.

\subsubsection{Occupational Employment Survey}

The {\em occupational employment survey} dataset \cite{spyromitros:2014:arxiv} was obtained from the annual occupational employment survey that is performed by the US Bureau of Labor Statistics. Every instance contains the aproximate number of full-time equivalent employees of different employment positions for a specific city. There are two versions of this datasets, one with data for 334 cities in the year 1997 (oes1997) and one with data for 403 cities in the year 2010 (oes2010). The employment types that were present in at least 50\% of the cities were considered as variables. From these, the targets are 16 randomly selected variables, while the rest constitute the input variables. 

\subsubsection{River Flow}

The {\em river flow} dataset \cite{spyromitros:2014:arxiv} was constructed for the prediction of the flow in a river network at 8 specific sites, 48 hours in the future. Those sites are located in the Mississippi River in the USA. There are two versions of this dataset. River Flow 1 (rf1) contains 64 input variables that refer to the most recent observations of the 8 sites and the observations from 6, 12, 18, 24, 36, 48 and 60 hours in the past. River Flow 2 (rf2) contains additional input variables that refer to precipitation forecasts for 6 hour windows up to 48 hours in the future for each gauge site. The target attributes are 8, each one corresponding to each of the 8 sites. The data were collected from September 2011 to September 2012.

\subsubsection{Solar Flare}

The {\em solar flare} dataset \cite{asuncion+newman:2007} has 3 target variables that correspond to the number of times 3 types of solar flare (common, moderate, severe) are observed within 24 hours. There are two versions of this dataset. Solar Flare 1 (sf1969) contains data from year 1969 and Solar Flare 2 (sf1978) from year 1978.

\subsubsection{Water Quality}

The {\em water quality} dataset \cite{dzeroski:2000} has 14 target attributes that refer to the relative representation of plant and animal species in Slovenian rivers and 16 input attributes that refer to physical and chemical water quality parameters.

\subsubsection{Supply Chain Management}

The {\em supply chain management} dataset \cite{spyromitros:2014:arxiv} is obtained from the Trading Agent Competition in Supply Chain Management (TAC SCM) tournament from 2010. The precise methods for data preprocessing and normalization are described in detail in \cite{groves+gini:2011}. Some benchmark values for prediction accuracy in this domain are available from the TAC SCM Prediction Challenge \cite{Pardoe10a}. These data sets correspond only to the {\em Product Future} prediction type. The input attributes contain the observed prices for a specific day in the tournament for each game. Moreover, 4 time-delayed observations for each observed product and component (1, 2, 4 and 8 days delayed). The target attributes are 16 and refer to the next day mean price (scm1d dataset) or the mean price within the next 20 days (scm20d dataset).

\section{Results}
\label{sec:results}

\subsection{Investigation of Parameters}

We first investigate the behaviour of our method with respect to its two main parameters: the number of models, $r$, which we vary from $q$ to 500 and the number of targets that are being combined, $k$, which we vary from 2 to $q$.

\begin{figure}
\centering
\includegraphics[width=\textwidth]{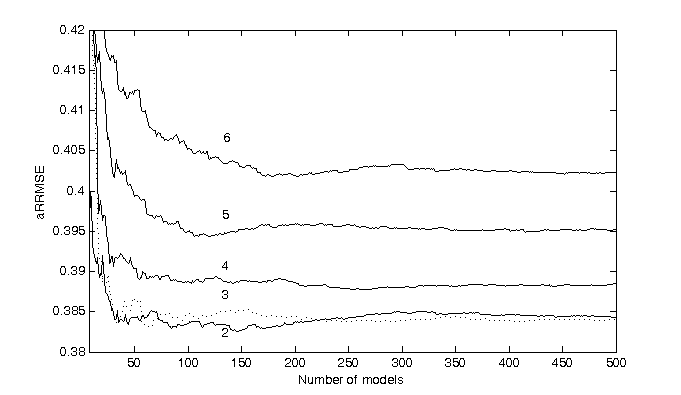}
\caption{\label{fig:atp1d}aRRMSE of our method (y-axis) for $k \in \{2,3,4,5,6\}$ with respect to the number of participating regression models (x-axis) at the {\em atp1d} dataset. The line corresponding to $k=3$ is dotted instead of solid, so as to contrast it with the overlapping line of $k=2$.}
\end{figure}

Figure \ref{fig:atp1d} shows the aRRMSE of our method (y-axis) at the {\em atp1d} dataset with respect to $r$ (x-axis) for $k \in \{2,3,4,5,6\}$. We notice that the curves have logarithmic shape, steeply decreasing with approximately the first 50 models and converging after approximately 250 models. The addition of models has the typical error-correction behaviour exhibited by ensemble methods, in accordance with our expectations. We further notice, again as we expected, that low numbers of $k$ (2 and 3) lead to the best results.

\begin{figure}
\centering
\includegraphics[width=\textwidth]{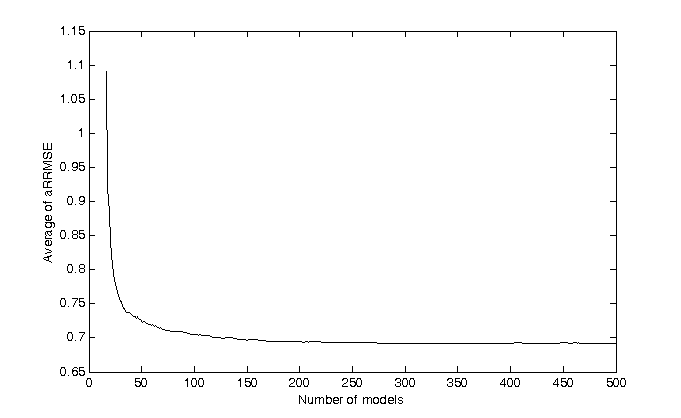}
\caption{\label{fig:meanPerModel}Average aRRMSE of our method (y-axis) with respect to $r$ (x-axis) across all datasets and all different $k$ values.}
\end{figure}

The behaviour of our approach with respect to $r$ is similar in all datasets. Figure \ref{fig:meanPerModel} shows the average aRRMSE of our method (y-axis) with respect to $r$ (x-axis) across all datasets and all $k$ values. Averages of performance estimates across datasets are not appropriate for summarizing and comparing the accuracy of different methods \cite{demsar:2006} and averages across different values of a parameter may hide salient effects of this parameter. However, we believe that this average serves well our purpose of summarizing a large number of results in a concise way in order to highlight the general behaviour of our method, which is consistent across all datasets and $k$ values. The number of participating models starts from 16, to ensure that the displayed average values are based on all datasets (recall that the minimum number of models in our approach is $q$ and that the maximum number of labels across our datasets is 16). We again see that the error follows the shape of a logarithmic curve, steeply decreasing with the first approximately 75 models and converging after approximately 280 models.

The performance of our approach with respect to $k$ is similar in all datasets too. The first 16 rows of Table \ref{tbl:per-target} shows the aRRMSE of our method for 500 models. We notice that the best results of our approach, which are underlined in the table, are obtained for $k \in \{2,3\}$, while the error is in most cases monotonically increasing with higher values of $k$.

\begin{table}
\begin{center}
{\caption{aRRMSE of our method in each dataset for $r=500$ and all possible $k$ values. The best result of our approach in each dataset is underlined. The last two rows show the aRRMSE of ST and MORF.}\label{tbl:per-target}}
\setlength{\tabcolsep}{2pt}
\begin{scriptsize}
\begin{tabular}{ccccccccccccc}
\hline\noalign{\smallskip}
$k$ 	 & atp1d       & atp7d       & edm         & sf1969  	  & sf1978      & oes10		& oes97		& rf1    		& rf2    		& scm1d & scm20d  & wq \\
\hline\noalign{\smallskip}
2        & 0.3842      & \underline{0.4614}& \underline{0.6996}& 1.2312 	  & 1.5746      & \underline{0.5026} 	& 0.5593 		& \underline{0.7265}  & \underline{0.7036}  & \underline{0.4572} & 0.7469 & 0.9100        \\
3        & \underline{0.3840}& 0.4653      &             & \underline{1.2172} & \underline{1.5675}& 0.5084  		& \underline{0.5588}  & 0.7878 		& 0.7584 		& 0.4610 & \underline{0.7467} & \underline{0.9080}        \\
4        & 0.3884      & 0.4796      &             &        	  &             & 0.5232 		& 0.5730 		& 0.8204 		& 0.7922 		& 0.4663 & 0.7472 & 0.9085        \\
5        & 0.3952      & 0.4917      &             &        	  &             & 0.5359 		& 0.5837 		& 0.8584 		& 0.8327 		& 0.4699 & 0.7477 & 0.9086        \\
6        & 0.4022      & 0.5029      &             &        	  &             & 0.5472 		& 0.5889 		& 0.8515 		& 0.8257 		& 0.4775 & 0.7490 & 0.9089        \\
7        &             &             &             &        	  &             & 0.5551 		& 0.5958 		& 0.8446 		& 0.8106 		& 0.4820 & 0.7513 & 0.9090        \\
8        &             &             &             &        	  &             & 0.5734 		& 0.6076 		& 0.8868 		& 0.8655 		& 0.4855 & 0.7536 & 0.9107        \\
9        &             &             &             &        	  &             & 0.5911 		& 0.6153 		&        		&        		& 0.4889 & 0.7548 & 0.9122        \\
10       &             &             &             &        	  &             & 0.6031 		& 0.6229 		&       		&        		& 0.4932 & 0.7537 & 0.9128        \\
11       &             &             &             &        	  &             & 0.6154 		& 0.6348 		&        		&        		& 0.4978 & 0.7573 & 0.9150        \\
12       &             &             &             &        	  &             & 0.6285 		& 0.6449 		&        		&        		& 0.5020 & 0.7571 & 0.9163        \\
13       &             &             &             &        	  &             & 0.6354 		& 0.6590 		&        		&        		& 0.5057 & 0.7619 & 0.9188        \\
14       &             &             &             &        	  &             & 0.6428 		& 0.6682 		&        		&        		& 0.5133 & 0.7640 & 0.9217        \\
15       &             &             &             &        	  &             & 0.6525 		& 0.6860 		&        		&        		& 0.5155 & 0.7681 &               \\
16       &             &             &             &        	  &             & 0.6652 		& 0.6916 		&        		&        		& 0.5218 & 0.7704 &               \\
\hline\noalign{\smallskip}
ST       & 0.3980 	   & 0.4735 	 & 0.7316 	   & 1.2777 	  & 1.6158      & 0.5421 		& 0.5727 		& 0.7171 		& 0.6897 & 0.4625 & 0.7571 & 0.9200        \\
\textsc{morf}     & 0.4223      & 0.5508 	 & 0.7338 	   & 1.2620 	  & 1.4020      & 0.4528 		& 0.5490 		& 0.8488 		& 0.9189 & 0.5635 & 0.7775 & 0.8994       \\
\hline
\end{tabular}
\end{scriptsize}
\end{center}
\end{table}

\subsection{Comparative Evaluation}

The last two rows of Table \ref{tbl:per-target} shows the aRRMSE of the ST strong baseline and the MORF state-of-the-art approach. To compare our approach with ST and MORF, we follow the recommendations of \cite{demsar:2006}. We first discuss the number of datasets where each of the methods is better than each of the others based on Table \ref{tbl:pairwise}. We see that RLC with $r=500$ is better than ST in 10/12 datasets and better than MORF in 8/12 datasets, both for $k=2$ and for $k=3$. The strength of the baseline is demonstrated by the fact that it is better than MORF in 7/12 datasets.

\begin{table}
\renewcommand{\tabcolsep}{0.15cm}
\begin{center}
{\caption{Number of datasets where a method is better than another method (wins:losses) for each pair of methods.}\label{tbl:pairwise}}
\begin{tabular}{cccc}
\hline\noalign{\smallskip}
 	 & RLC & ST & MORF   \\
\hline\noalign{\smallskip}
RLC  &  -  & 10:2 & 8:4 \\
ST   & 2:10&  -   & 7:5 \\
MORF & 4:8 & 5:7  & - \\
\noalign{\smallskip}\hline
\end{tabular}
\end{center}
\end{table}

The mean rank of RLC with $r=500$ and $k=2$ or $k=3$ (same $k$ for all datasets), ST and MORF are 1.5, 2.25 and 2.25 respectively. The variation of the Friedman test described in \cite{demsar:2006} to compare the three algorithms rejects the null hypothesis for a $p$-value of 0.0828 (i.e. requires $a=0.1$). Proceeding to a post-hoc Nemenyi test with $a=0.1$, the critical difference is 0.8377, slightly more than the 0.75 difference among the mean rank of RLC and that of ST and MORF. So, these differences should not be considered statistically significant based on this test.

We also applied the Wilcoxon signed-ranks test between RLC with $r=500$ and $k=2$ and the other two algorithms. While multiple tests are involved in this process, these are limited to just 2, and therefore a small bias will be introduced if any due to this multiple testing process. For the comparison with ST the $p$-value is 0.0210 suggesting that the differences are statistically significant for $a=0.05$, while for the comparison with MORF the $p$-value is 0.1763 suggesting that the differences are statistically insignificant even for $a=0.1$.

One could argue that a fairer comparison between RLC and MORF should have setup MORF to use 500 trees instead of 100. The answer to such critique is that each target is involved in $rk$/$q$ regression models in RLC and thus in datasets such as {\em oes}, {\em scm} and {\em wq}, RLC is actually at disadvantage. Three of the wins of MORF over RLC actually occur in the {\em oes} and {\em wq} datasets. Perhaps a fairer experiment would set $r=100q/k$, assuming 100 trees in MORF. Selecting the number of models in RLC and MORF via cross-validation would perhaps be even fairer. Such experiments will be considered in future work.

Summarizing the comparative results, we argue that the proposed approach is worthy of being considered by a practitioner for a multi-target regression domain, as there is a high chance that it could give the best results compared to state-of-the-art methods. Futhermore, being algorithm independent, it has the flexibility and potential of doing better in a specific application, by being instantiated with a different base learner whose hypothesis representation is more suited to the given problem (e.g. a support vector regression algorithm), in contrast to MORF (and other variants of the predictive clustering trees framework), whose representation is fixed to trees.

\subsection{Error with Respect to Average Pairwise Target Correlation}

No clear conclusion can be drawn on whether the intensity of pairwise correlations affects the improvement that our approach can give over the baseline. The correlation among the median of the absolute value of pairwise target correlations and the gain in performance over ST is 0.15.

Noticing that the high variance of pairwise correlations in the river-flow datasets co-occurs with the failure of our approach to improve upon ST, we also calculated the correlation between the standard deviation of the pairwise target correlations and the gain in performance over ST, which is -0.68 ({\em edm} was excluded in this computation as it only has two targets). This apparently suggests that low variance of absolute value of pairwise target correlations leads to improved gains. However, we do not have a theory to explain this correlation.

Pairwise target correlations do not take the input features into account, so they do not measure potential conditional dependencies among targets given the inputs \cite{dembczynski:2010c}. We do however notice that in the three pairs of datasets with similar nature and amount of features (the two versions of {\em atp}, {\em oes} and {\em sf} datasets), higher median of absolute value of pairwise target correlations does lead to improved performance. We simplistically assume here that similar nature and amount of features introduce similar conditional dependencies of the targets given the features, even though the aforementioned pairs of datasets have different, yet of similar nature, targets.

Table \ref{tbl:corr} presents the data, upon which the discussion of this subsection is based. In specific, the 1st row shows the percentage of improvement of our approach compared to ST, while the next two rows show the median and standard deviation respectively of the absolute value of pairwise target correlations.

\begin{table}
\begin{center}
{\caption{For each dataset, the 1st row shows the percentage of accuracy gain of our method compared to ST, and the next two rows show the median and standard deviation respectively of the absolute value of pairwise target correlations.}\label{tbl:corr}}
\setlength{\tabcolsep}{1.4pt}
\begin{scriptsize}
\begin{tabular}{ccccccccccccc}
\hline\noalign{\smallskip}
 	 	 & atp1d  & atp7d     & edm   & sf1969  & sf1978   & oes10	& oes97		& rf1   	& rf2    & scm1d  & scm20d & wq \\
\hline\noalign{\smallskip}
gain (\%)& 3.6    & 2.6 	 & 4.6 	  & 5.0 	& 3.1      & 7.9 	& 2.5 		& -1.3 		& -2.0   & 1.6    & 1.4    & 1.3     \\
median 	 & 0.8013 & 0.6306 	 & 0.0051 & 0.2242 	& 0.1484   & 0.8479 & 0.7952 	& 0.4077	& 0.4077 & 0.6526 & 0.5785 & 0.0751  \\
stdev 	 & 0.0788 & 0.1602 	 & - 	  & 1.1247 	& 1.2006   & 0.0972 & 0.0785 	& 0.3125	& 0.3125 & 0.1316 & 0.1483 & 0.0717  \\
\hline
\end{tabular}
\end{scriptsize}
\end{center}
\end{table}

To the best of our knowledge, a discussion of accuracy with respect to target dependencies has not been attempted in past multi-target regression work. We believe such an analysis is quite interesting both theoretically and practically and might be good on one hand to be adopted by future work in this area, and on another hand to be studied more elaboratively by itself.

\section{Conclusions and Future Work}
\label{sec:conclusions}

Multi-target regression is a learning task with interesting practical applications. We expect its popularity to rise in the near future with the proliferation of multiple sensors in our everyday life (Internet of Things) recording multiple values that we might want to predict simultanteously.

Motivated from the practical interest of multi-target regression and recent work on drawing parallels between multi-label classification and multi-target regression, we developed an ensemble method that constructs new target variables by forming random linear combinations of existing targets, as a twin of the RA$k$EL multi-label classification algorithm. At the same time, we highlighted an additional connection of the proposed approach with recent multi-label classification algorithms based on output coding.

The proposed approach was found significantly better than a strong baseline that learns a single model per target using gradient boosting and compares favourably against the state-of-the-art ensemble method MORF, based on experiments on 12 multi-target regression datasets. Furthermore, the empirical study reveals a relation among the pairwise correlation of targets and the gains of the proposed approach given similar input features, suggesting succesful exploitation of existing unconditional target dependencies by the proposed approach.

The proposed approach has the potential to be further improved in the future. Towards that direction, we intend to investigate alternative randomization injection processes (e.g. normal instead of uniform coefficients) and constructing ensembles of our approach using different coefficient matrices. For example, instead of constructing 500 models with one matrix, we could construct 100 models with 5 different matrices, which is expected to improve diversity and potentially accuracy of our idea.

\subsubsection{Acknowledgements.} This work has been partially supported by the Greek General Secretariat for Research and Technology, via act {\em Supporting Groups of Small and Medium-Sized Enterprises for Resarch and Technological Development Activities}, project 22SMEs2010, {\em Intelligent System in Supply Chain Monitoring and Optimization}.

\bibliographystyle{splncs}
\bibliography{all}
\end{document}